\documentclass[letterpaper, 10 pt, conference]{ieeeconf}  

\IEEEoverridecommandlockouts                              

\usepackage{graphics} 
\usepackage{epsfig} 
\usepackage{amsmath} 
\usepackage{amssymb}  
\usepackage{subcaption}
\usepackage{graphicx}
\usepackage{adjustbox}
\usepackage{multirow}
\usepackage{makecell}
\usepackage{hyperref}

\title{\LARGE \bf
Follow the Footprints: Self-supervised Traversability Estimation for Off-road Vehicle Navigation based on Geometric and Visual Cues
}

\author{Yurim Jeon, E In Son and Seung-Woo Seo \\
Seoul National University, Seoul, Republic of Korea \\
\tt\small \{fabioisyo01, pingpang, sseo\}@snu.ac.kr
\thanks{This work was supported by Institute of Information \& communications Technology Planning \& Evaluation (IITP) grant funded by 
the Korea government (MSIT) (No. RS-2022-00207391, Development of Hashgraph-based Blockchain Enhancement Scheme and Implementation 
of Testbed for Autonomous Driving)}%
}

\begin{document}

\maketitle
\thispagestyle{empty}
\pagestyle{empty}

\begin{abstract}

In this study, we address the off-road traversability estimation problem, that predicts areas where a robot can navigate in off-road environments. An off-road environment is an unstructured environment comprising a combination of traversable and non-traversable spaces, which presents a challenge for estimating traversability. This study highlights three primary factors that affect a robot's traversability in an off-road environment: surface slope, semantic information, and robot platform. We present two strategies for estimating traversability, using a guide filter network (GFN) and footprint supervision module (FSM). The first strategy involves building a novel GFN using a newly designed guide filter layer. The GFN interprets the surface and semantic information from the input data and integrates them to extract features optimized for traversability estimation. The second strategy involves developing an FSM, which is a self-supervision module that utilizes the path traversed by the robot in pre-driving, also known as a footprint. This enables the prediction of traversability that reflects the characteristics of the robot platform. Based on these two strategies, the proposed method overcomes the limitations of existing methods, which require laborious human supervision and lack scalability. Extensive experiments in diverse conditions, including automobiles and unmanned ground vehicles, herbfields, woodlands, and farmlands, demonstrate that the proposed method is compatible for various robot platforms and adaptable to a range of terrains. Code is available at \url{https://github.com/yurimjeon1892/FtFoot}.

\end{abstract}

\section{INTRODUCTION}

Traversability estimation is the task of predicting the terrain that a ground robot can traverse using data obtained from sensors. The most typically used sensors for this task are cameras and LiDAR, which provide two-dimensional images and three-dimensional (3D) point clouds, respectively. The prediction output is a traversability map, which is used to plan the robot's driving path. As such, this task is crucial in ensuring the safe navigation of the robot. Failure of the traversability estimation may cause the robot to traverse the wrong path, thus potentially resulting in hazardous situations such as derailment, collision, or overturning. In particular, traversability estimation in off-road environments presents unique challenges that are not encountered in structured environments such as urban or indoor environments. These challenges include scattered obstacles and difficulty in distinguishing between traversable and non-traversable spaces in off-road terrains.

\begin{figure}[t]
    \begin{center}
    \includegraphics[width=1.\linewidth]{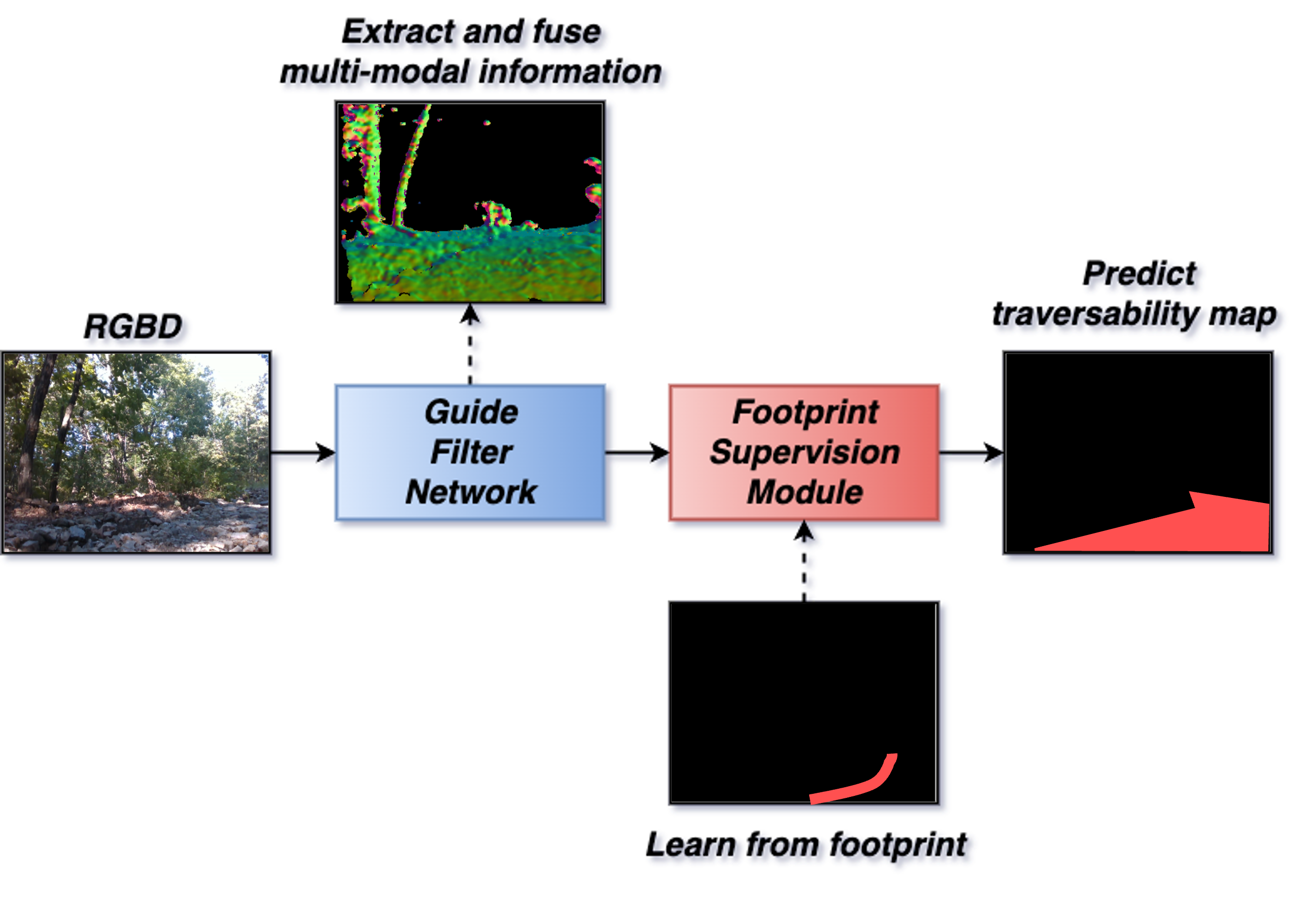}
    \end{center}
    \vspace*{-0.5cm}
    \caption{\textbf{Overview of the proposed method} 
    The proposed method consists of two components: a guide filter network optimized for feature extraction and fusion in off-road traversability estimation, and a footprint supervision module that learns traversability in a self-supervised manner from the robot's path, referred to as the footprint. 
    }
    \label{fig:itr_overview}
\end{figure}

Efforts have been expended continuously to estimate traversability in off-road environments. The first approach explicitly defines traversable spaces as human-supervised labels and employs rule-based~\cite{howard2000real}\cite{howard2001rule}\cite{kim2007traversability}\cite{talukder2002fast} or learning-based~\cite{kragh2015object}\cite{maturana2018real} methods to predict them. The second approach is a self-supervised approach, that combines commonly used exteroceptive sensors (e.g., camera and LiDAR) with newly added proprioceptive sensors (e.g., IMU)~\cite{wellhausen2019should}\cite{waibel2022rough}\cite{castro2022does}\cite{sathyamoorthy2022terrapn}\cite{gasparino2022wayfast}. The first approach is intuitive, but it requires laborious human supervision and fails to consider the robot platform in estimation. The second approach relies heavily on the proprioceptive sensor, which is sensitive to changes in the robot platform. Therefore, the traversability estimated by this method is applicable to only a few robot platforms and thus not scalable to other robot platforms.

The traversability of off-road terrains is not solely determined by obstacles or ground appearance. It is affected by multiple factors involving the off-road environment and robot platform. This study highlights three primary factors that affect a robot's traversability in an off-road environment: surface slope (geometric cue), semantic information (visual cue), and robot platform. These three factors determine the traversability through interactions; In cases where the surface slope of the terrain sharply increases, the area acts as an obstacle and becomes non-traversable. However, if the area's semantics can be penetrated, such as reeds, it becomes traversable. Lastly, even semantically non-traversable areas like rubble may still vary in traversability depending on the robot platform. For example, all-terrain vehicles can traverse over rubble, whereas small unmanned ground vehicles (UGVs) cannot do the same. Previous studies failed to predict robot-dependent traversability, by considering only geometric obstacles~\cite{talukder2002fast} or semantic information~\cite{maturana2018real}. Studies focused on a robot platform have limitations in compatibility with different robot platforms~\cite{wellhausen2019should}\cite{sathyamoorthy2022terrapn}.

We introduce two strategies to consider all three factors in traversability estimation. The first involves designing a neural network named as a \textbf{guide filter network (GFN)}, which extracts \textit{surface} and \textit{semantic information} from exteroceptive sensor data and integrates them to produce features optimized for traversability estimation. This is achieved by implementing an extraction network that extracts surface normals and semantic information from the input data, along with a fusion network composed of newly proposed guide filter layers, which offer advantages in the fusion of information from different modalities. The second strategy involves a \textbf{footprint supervision module (FSM)}, i.e., a self-supervision module that utilizes the footprint, which is the path traversed by the robot in pre-driving and thus incorporates the characteristics of the \textit{robot platform}. This module enables a more scalable traversability estimation by providing a common traversability score that is compatible with diverse robot platforms and does not require laborious human supervision. Using these two strategies, we developed a self-supervised traversability estimation method applicable to various robot platforms based on the geometric and visual information of the surrounding environment. An overview of the proposed method is shown in Fig.~\ref{fig:itr_overview}.

We demonstrate the effectiveness of our method through experiments using the public off-road dataset RELLIS-3D~\cite{jiang2020rellis3d} and ORFD~\cite{min2022orfd}. Our method successfully estimates the traversable space under various conditions, including different types of terrain, weather, lighting and robot platforms. The results show that the proposed method is a viable and optimized solution for estimating traversability in off-road environments.

The key contributions of this study are as follows:
\begin{itemize}
	\item A novel approach for traversability estimation is proposed that considers critical factors that determine off-road traversability: surface slope, semantic information, and robot platform.
	\item A new network named the guide filter network, which is optimized for traversability estimation, is designed; it offers advantages in extracting and fusing information from different modalities.
	\item A footprint supervision module that can predict the robot-dependent traversability based on the robot's footprint in a self-supervised manner is developed to improve the versatility of the algorithm.
	\item The performance of the proposed method is demonstrated through experiments using datasets from various terrains, weather, lighting and robot platforms.
\end{itemize}

\section{RELATED WORK}

Early traversability estimation studies primarily focused on urban and indoor environments. In these structured environments, the definition of traversable space is clear. Therefore, the approach of predefining the traversable space for estimation demonstrated desirable performance. In an urban environment, a traversable space was typically defined as a free space, which is an empty space on a paved road without obstacles~\cite{yao2015estimating}\cite{tsutsui2018minimizing}. Similarly, in indoor environments, the presence of obstacles was a significant factor that determines traversability~\cite{hirose2018gonet}\cite{hirose2019vunet}.

\begin{figure*}[ht!]
    \begin{center}
    \includegraphics[width=1.\linewidth]{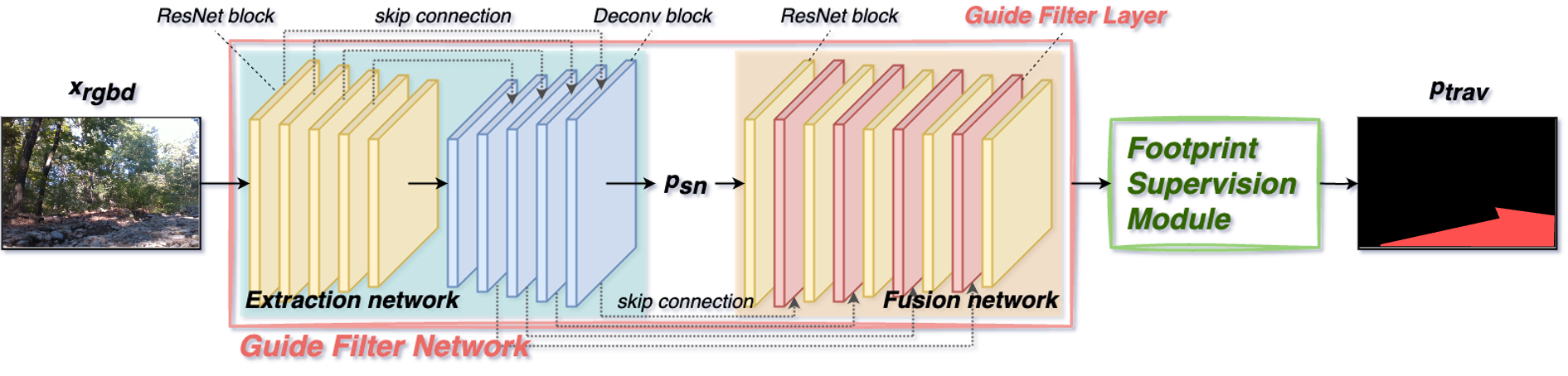}
    \end{center}
    \vspace*{-0.5cm}
    \caption{\textbf{Structure of the proposed method} 
    The guide filter network (GFN) consists of an extraction network and a fusion network. The extraction network estimates the surface normal image $p_{sn}$ from the input RGB-D image $x_{rgbd}$. Subsequently, a fusion network composed of guide filter layers integrates information from different modalities. Finally, the footprint supervision module (FSM) predicts the traversability map $p_{trav}$.
    }
    \label{fig:mtd_gfn}
\end{figure*}

Early studies on off-road conditions used the same approach of predefining traversability. Rule-based methods extracted physical features such as roughness, slope, discontinuity, and hardness from input data to classify traversable spaces~\cite{howard2000real}\cite{howard2001rule}. Another method used appearance features from the superpixels of an RGB image for classification~\cite{kim2007traversability}. Additionally, there is a method that pre-defined the characteristics of off-road obstacles in 3D data and segmented obstacles based on these characteristics~\cite{talukder2002fast}. A support vector machine-based method estimated traversability by classifying 3D point clouds into ground, vegetation, or objects using SVM~\cite{kragh2015object}. A semantic segmentation-based method predicted semantic classes from RGB images, such as road, grass, tree, and obstacle, and estimated traversability based on these classes~\cite{maturana2018real}. 
However, these methods required laborious human supervision and cannot predict traversability varies depending on the robot platform.

Subsequently, a self-supervised approach using proprioceptive sensor data from a robot’s driving experience was introduced. There is a method that employed self-supervision, utilizing self-generated ground-truth cost derived from force-torque signals in a proprioceptive sensor to train the neural network model~\cite{wellhausen2019should}. Other methods utilized self-supervision from IMU data~\cite{waibel2022rough} and the robot's velocity~\cite{castro2022does}\cite{sathyamoorthy2022terrapn} to estimate traversability. Overall, these methods, use the robot's state as self-generated ground truth, making them incompatible with various robot platforms. Lastly, WayFAST~\cite{gasparino2022wayfast}, the closest method to ours, integrated RGB and depth images using the standard convolution layer. It used a narrow area around the robot's trajectory as self-generated ground truth. However, this method inefficiently integrated multi-modal information due to a standard convolution layer~\cite{jia2016dynamic} and had a limitation of providing diverse path planning options due to a narrow ground truth space.

Herein, we introduce a new traversability estimation method that addresses the limitations of the previous methods. The proposed method can predict traversability in a self-supervised manner and does not require human supervision. Furthermore, our method considers the robot platform in traversability estimation while also providing universally applicable traversability results compatible with various robot platforms.

\section{METHODS}

\subsection{Overview}

The structure of the proposed method is shown in Fig.~\ref{fig:mtd_gfn}. The network uses an RGB-D image denoted by $x_{rgbd}$ as the input. The input undergoes a sequence of ResNet~\cite{he2016deep} and deconvolution blocks. A skip connection is applied between ResNet block and deconvolution block to allow information flow between different layers, except for the first deconvolution block. In the skip connection, the output of the ResNet block is concatenated with the output of the previous deconvolution block and then input into the next deconvolution block. The output of the final deconvolution block is the surface normal image, denoted as $p_{sn}$. The surface normal image has a 3D surface normal value for each pixel in the image. The mean-squared error is used as the loss function for $p_{sn}$. A network from $x_{rgbd}$ to $p_{sn}$ is referred to as the extraction network.

After the extraction network, $p_{sn}$ is passed through a sequence of ResNet blocks and guide filter layers. The guide filter layer requires two inputs: the output of the previous ResNet block, which serves as a guidance image containing geometric information, and the output of the deconvolution block, which serves as a convolve image containing visual information. Through the guide filter layer, the network simultaneously considers both the surface slope contained in the geometric information as well as semantic information contained in the visual information. This network is referred to as the fusion network.

The output feature map is input into the footprint supervision module. The footprint supervision module is designed to estimate the entire traversable space using the footprint, which is a partial annotation of the entire traversable space. Consequently, a traversability map,~$p_{trav}$, is generated as the final output.

\subsection{Guide Filter Network (GFN)}

Traversability estimation requires the consideration of geometric (e.g., surface slope) and visual cues (e.g., semantic information). The challenge lies in extracting appropriate features by integrating information from different modalities. A common approach for this integration is using a neural network with standard convolutional layers. However, standard convolutional layers apply the same content-agnostic convolution filter to each image pixel, which leads to the limited feature representation~\cite{jia2016dynamic}\cite{zhou2021decoupled}. This limitation becomes critical in off-road environments, where traversability depends on the interaction between geometric and visual cues.

Thus, we devised a new layer, named the guide filter layer, which can extract features via different filters for each location and content by simultaneously considering different modalities. The guide filter layer generates different input-conditioned filters for each pixel in an image. These spatially variant and content-dependent filters can improve the feature representation.

The remainder of this section is organized as follows. Section~\ref{sec:gfn1} discusses the dynamic filter layer, which is the base of the guide filter layer. Section~\ref{sec:gfn2} discusses the structure of the guide filter layer.

\subsubsection{Background: Dynamic filter layer} \label{sec:gfn1}

The convolution operation of a dynamic filter layer can be formulated as follows:
\begin{equation}
	F^{*} (i,j) = K_{ij} \otimes F(i,j),
\end{equation}
where $F \in \mathbb{R}^{c \times h \times w}$ is the input feature map and $F^{*} \in \mathbb{R}^{c^{*} \times h^{*} \times w^{*}}$ is the output feature map. $K_{ij} \in \mathbb{R}^{c^{*} \times c \times k \times k }$ is a generated filter of size $k$ for pixel $(i,j)$ and $\otimes$ denotes the convolution operation. The filter~$K_{ij}$ can be decomposed~\cite{tang2020learning} to reduce the number of parameters per filter as follows:
\begin{equation}
\begin{aligned}
	F'(i,j) = K'_{ij} \otimes F(i,j) \\
	F''(i,j) = K''_{ij} \otimes F'(i,j),
	\label{eqn:ddf}
\end{aligned}
\end{equation}
where $F' \in \mathbb{R}^{c \times h' \times w'}$ is the intermediate feature map obtained by applying the spatially variant filter $K'_{ij} \in \mathbb{R}^{1 \times 1 \times k \times k}$ to $F$. The final feature map $F'' \in \mathbb{R}^{c' \times h' \times w'}$ is obtained by applying content-dependent filter $K''_{ij} \in \mathbb{R}^{c' \times c \times 1 \times 1}$ to $F’$.

\subsubsection{Guide filter layer} \label{sec:gfn2}

\begin{figure}[t]
    \begin{center}
    \includegraphics[width=1.\linewidth]{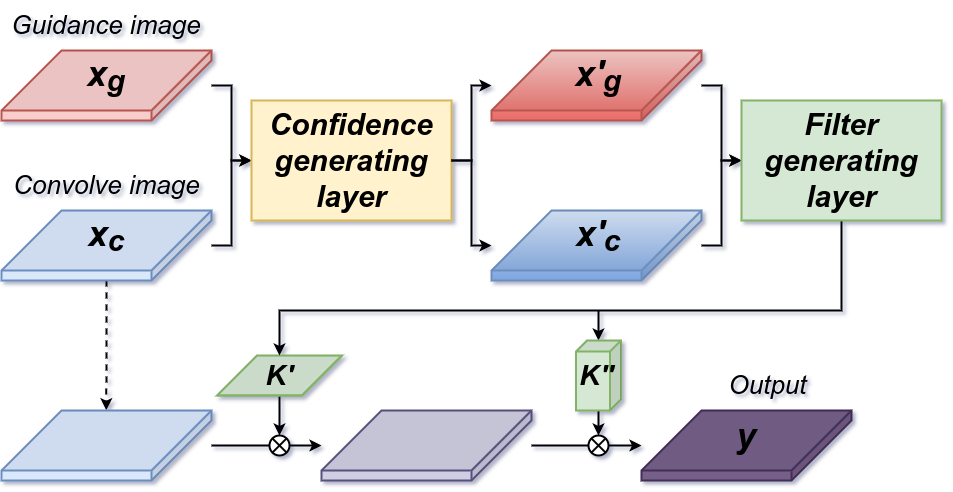}
    \end{center}
    \vspace*{-0.3cm}
    \caption{\textbf{Structure of the guide filter layer} 
    The guide filter layer takes two inputs: the guidance image $x_{g}$ and the convolve image $x_{c}$. First, these images are fed into a confidence-generating layer, which produces weighted images $x'_{g}$ and $x'_{c}$. These are then forwarded to the filter-generating layer to generate two decomposed filters, $K'$ and $K''$. Finally, these filters are sequentially applied to the initial input $x_{c}$, resulting in the output $y$.
    }
    \label{fig:mtd_gfl}
\end{figure}

We developed a new layer, named the guide filter layer, which utilizes the basic structure of the dynamic filter layer for the effective fusion of different modalities. This layer requires two inputs, i.e., a guidance image and a convolve image, which represents different modalities. A guidance image is primarily used to generate a dynamic filter, which represents a surface normal image containing geometric information. A convolve image, which is an RGB image containing visual information, is then convolved with the dynamic filter for feature extraction. The guide filter layer regulates the effect of the guidance and convolve images on filter generation, thus allowing it to extract task-optimized features by considering both images. This approach allows the guide filter layer to efficiently integrate information from different modalities. The structure of the guide filter layer is shown in Fig.~\ref{fig:mtd_gfl}.

Two inputs of the guide filter layer, i.e., the guidance image $x_{g} \in \mathbb{R}^{c \times h \times w}$ and convolve image $x_{c} \in \mathbb{R}^{c \times h \times w}$, are passed to the confidence-generating layer. The confidence-generating layer comprises a standard convolutional layer and a softmax layer. As output, a confidence score $w_{conf} \in \mathbb{R}^{2 \times h \times w}$ is generated. The two inputs, $x_{g}$ and $x_{c}$, are multiplied elementwise by $w_{conf}$ to obtain weighted inputs $x'_{g}$ and $x'_{c}$, respectively.

The weighted guidance image $x'_{g}$ and convolve image $x'_{c}$ are input to the filter-generating layer, which produces two types of filters per image pixel. The first filter is a spatially variant filter $K'_{ij} \in \mathbb{R}^{1 \times 1 \times k \times k}$, and the second is a content-dependent filter $K''_{ij} \in \mathbb{R}^{c' \times c \times 1 \times 1} $. The two generated filters are used sequentially for the convolution operation with $x_{c}$ based on Eq.~\ref{eqn:ddf}.

\subsection{Footprint Supervision Module (FSM)} 

Traversable space varies depending on the robot platform. Therefore, a recent approach utilizes a proprioceptive sensor, which measures the state of the robot, to train algorithms that can estimate robot-dependent traversability. However, the proprioceptive sensor is sensitive to minor changes in the robot platform, such as weight or sensor configurations. Thus, this approach makes the algorithm vulnerable to variations of the robot platform, hindering the scalability of the algorithm.

Our approach leverages the robot's trajectory in pre-driving as a footprint to capture the robot’s response to the terrain. This enables us to predict the robot-dependent traversability, which reflects the characteristics of the robot platform. Additionally, this approach resolves the compatibility issue by estimating a common traversability score that can be applied to various robot platforms. This idea is implemented under the name, footprint supervision module (FSM). The footprint provides a partial annotation of the entire traversable space, and the footprint supervision module propagates information to neighboring areas with similar characteristics, thus enabling the estimation of the complete traversable space from partial annotation. The structure of this module is designed based on scribble-supervised semantic segmentation~\cite{pan2021scribble}.

\begin{figure}[t!]
    \begin{center}
    \includegraphics[width=1.0\linewidth]{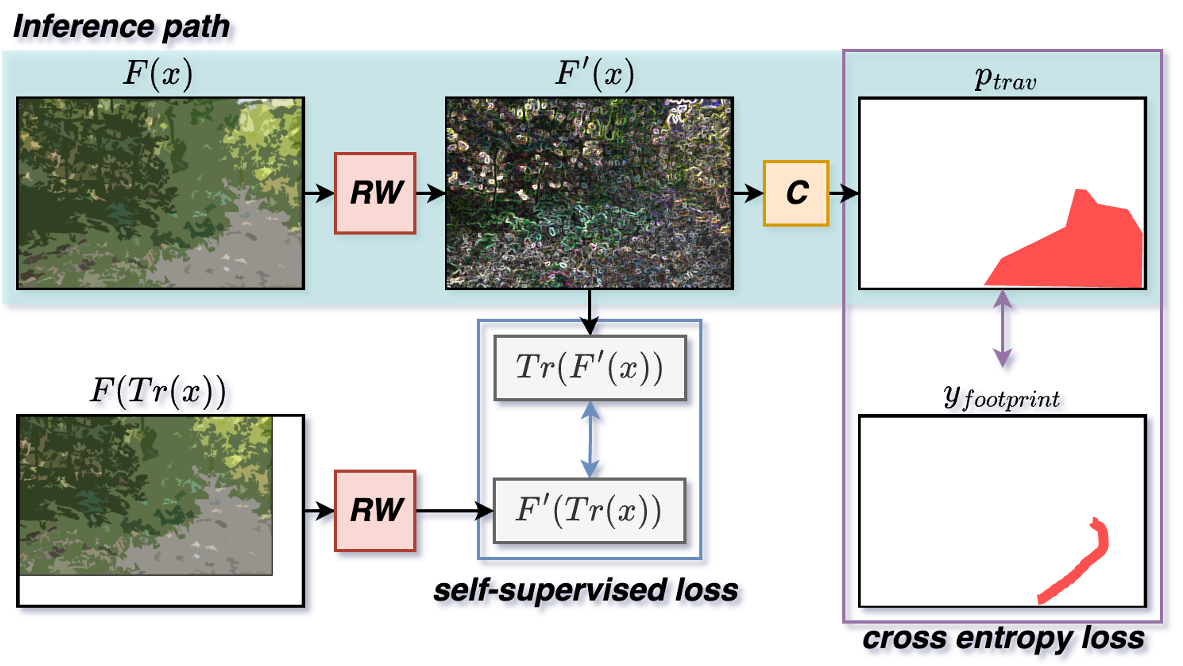}
    \end{center}
    \vspace*{-0.3cm}
    \caption{\textbf{Structure of the footprint supervision module} 
    The footprint supervision module takes an input feature map, $F(x)$. 
    This module consists of a random walk (RW) and a convolutional layer (C). $Tr$ represents the transformation function. $p_{trav}$ is the final traversability map. The inference path is represented by the light blue box.
    }
    \label{fig:mtd_fsm}
\end{figure}

Sections~\ref{sec:fsm1} and \ref{sec:fsm2} discuss the elements of the footprint supervision module. The structure of the footprint supervision module is illustrated in Fig.~\ref{fig:mtd_fsm}.

\subsubsection{Random walk} \label{sec:fsm1}

The first element of the footprint supervision module is a random walk that propagates information around the footprint. The random walk has been used in graph theory~\cite{lovasz1993random}. In a graph $G$ with $n$ nodes, an affinity matrix $A \in \mathbb{R}^{n \times n}$ can be defined, where $A_{ij}$ represents the affinity between nodes $i$ and $j$. Information propagation from a specific node within $G$ can be simulated using $A$, and the diffusion of information over time can be modeled through repeated matrix multiplication.

The application of the random-walk operation to a neural network can be expressed as follows:
\begin{equation}
F’ =\alpha AF+ F,   
\label{eqn:rw}
\end{equation}
where $F$ denotes the input feature map and $F’$ denotes the output. The affinity matrix $A$ indicates the probability of a random walk and is defined as $A = softmax(F^T F)$. Finally, $\alpha$ represents a learnable parameter that controls the degree of the random walk, which is the degree of information diffusion.

\subsubsection{Loss function} \label{sec:fsm2}

\textbf{Self-supervised loss}. Self-supervised loss is widely used in unsupervised learning to yield consistent network outputs from similar features. It is defined as the difference between the transformation of the neural representation generated from input $x$ and the neural representation generated from the transformation of $x$, and is expressed as follows:

\begin{equation}
L_{ss} = dist(Tr(F(x)), F(Tr(x))),
\end{equation}
where $F(x)$ is the feature map from input $x$, $Tr$ is the transformation function, and $dist$ is the distance metric between the two neural representations. In this study, $Tr$ is used as the horizontal flip and translation, and $dist$ is used as the mean squared error.

\textbf{Cross entropy loss}. The cross-entropy loss is calculated between the robot's footprint~$y_{footprint}$ and predicted traversability map $p_{trav}$. Because the footprint represents only a partial area of the entire traversable space, the negative labels in $y_{footprint}$ include both positive and negative samples of the traversable space. Therefore, we assign weights of $1$ and $0.1$ to the losses on positive and negative labels, respectively.

\section{EXPERIMENTS}

\subsection{Dataset}

\begin{figure}[t!]
    \begin{center}
    \includegraphics[width=1.\linewidth]{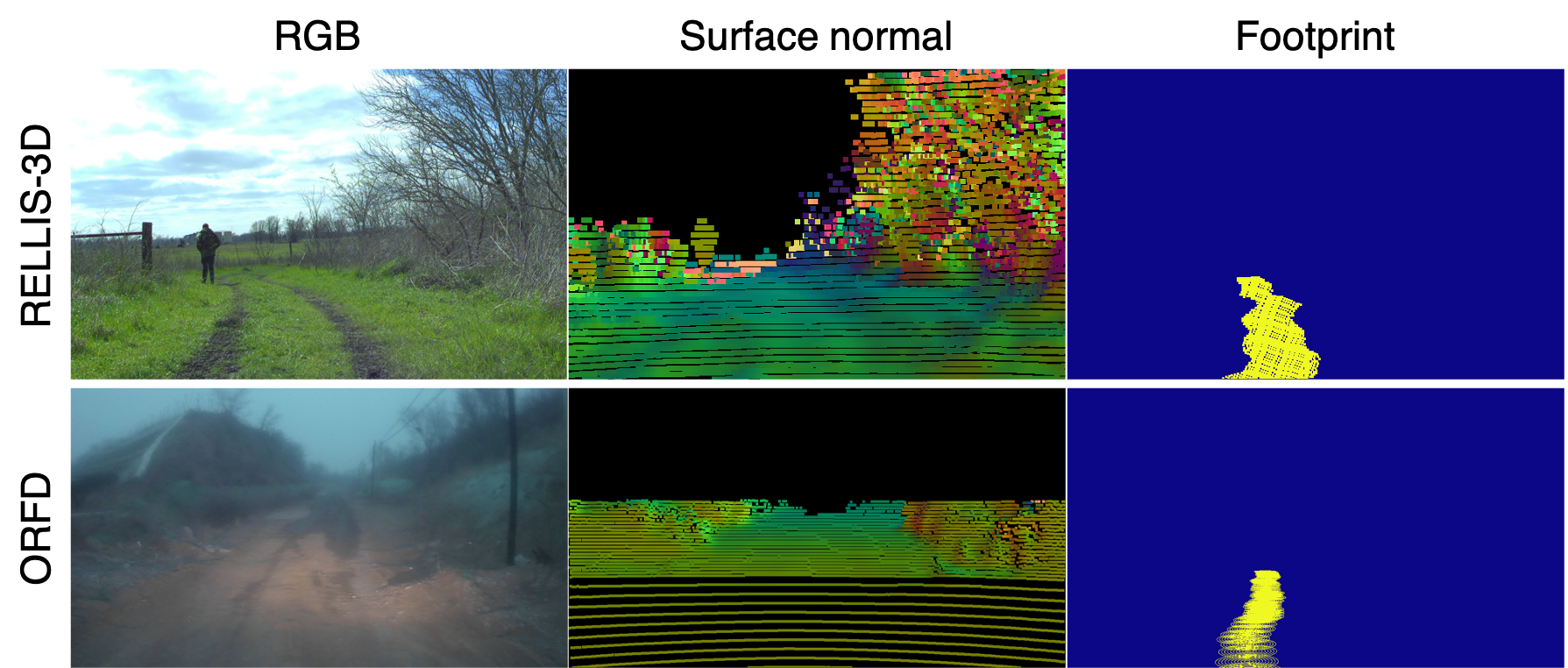}
    \end{center}
    \vspace{-0.3cm}
    \caption{\textbf{Examples of the dataset} 
    Sample images from the RELLIS-3D and ORFD datasets.
    }
    \label{fig:exp_sample}
\end{figure}

\begin{figure*}[ht!]
    \begin{center}
    \includegraphics[width=0.8\linewidth]{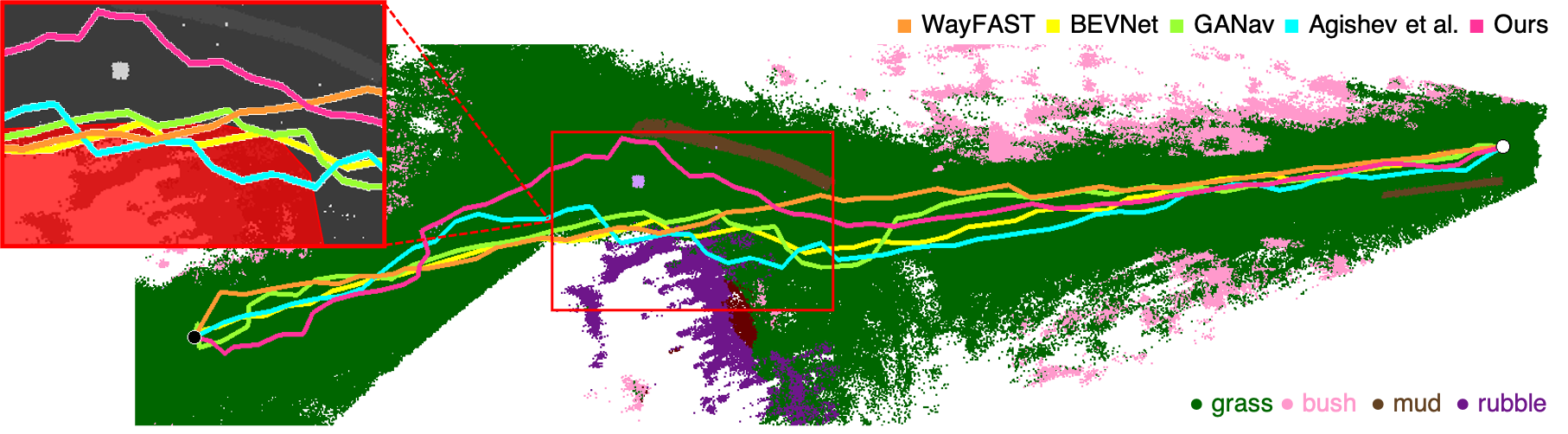}
    \end{center}
    \vspace*{-0.4cm}
    \caption{\textbf{Path generated on the RELLIS-3D test route 1}  
    The red area indicates dangerous terrain for driving due to rocks. \textit{Ours} (pink line) is the only safe path that avoids this area.
    }
    \label{fig:exp_rellis_path}
\end{figure*}

The RELLIS-3D~\cite{jiang2020rellis3d} dataset is an off-road dataset obtained using the Clearpath Robotics Warthog UGV platform. It provides RGB images captured using a Basler acA1920-50gc camera, point clouds from the Ouster OS1, and pose data from GPS. The dataset contains four data sequences, each comprising more than 2000 samples. We used sequences 0, 1, 2, and 3 for training (11497 samples) and sequence 4 for testing. Sequence 4 was split into three test routes, each containing 500, 500, and 700 frames, respectively.

The ORFD~\cite{min2022orfd} dataset is another off-road dataset collected using the FAW Mazda Ruiyi 6 Coupe automobile platform. It includes RGB images and point clouds obtained using the Hesai Pandora all-in-one sensor kit. The dataset covers diverse terrains (woodland, farmland, and grassland), weather conditions (sunny, foggy, and snowy), and lighting variations (daylight, darkness, etc.). It provides 8392 samples for training, 1245 for validation, and 2193 for testing.

The footprint of the robot is generated based on the robot's pose data. First, the robot's trajectory is generated from the pose data, and projected onto the image plane. Next, the projected trajectory is transformed into a binary image, where areas the robot traversed are marked as 1, and others as 0.
 The example of the dataset is shown in Fig.~\ref{fig:exp_sample}.

\begin{table*}[t!]
\caption{Quantitative comparison of path planning on the RELLIS-3D dataset}
\centering
\begin{adjustbox}{width=1\linewidth}
\begin{tabular}{|c|ccc|ccc|ccc|c|}
\hline
                & \multicolumn{3}{c|}{Route 1}                  & \multicolumn{3}{c|}{Route 2}                & \multicolumn{3}{c|}{Route 3} &              \\ \cline{2-10}
Method          & SR           & CTE (m)         & HD (m)       & SR         & CTE (m)        & HD (m)        & SR         & CTE (m)        & HD (m)        & Runtime (ms) \\ \hline \hline
WayFAST~\cite{gasparino2022wayfast}        
                & 0.8          & 0.089          & 2.23          & \textbf{1} & 0.085          & 1.99          & \textbf{1} & 0.063          & 1.54          & 6.51   \\
BEVNet~\cite{shaban2022semantic}         
                & 0.63         & 0.116          & 3.04          & \textbf{1} & 0.093          & 3.13          & 0.7        & 0.137          & 2.32          & 86.31  \\
GANav~\cite{guan2022ga}          
                & 0.57         & 0.113          & 3.57          & -          & -              & -             & -          & -              & -             & 34.51  \\
Agishev et al.~\cite{agishev2022trajectory}
                & 0.5          & 0.131          & 2.43          & 0.67       & 0.134          & 2.39          & 0.87       & 0.113          & 2.73          & 60.17  \\
Ours           
                & \textbf{1}   & \textbf{0.066} & \textbf{2.08} & \textbf{1} & \textbf{0.071} & \textbf{1.55} & \textbf{1} & \textbf{0.061} & \textbf{1.18} & 16.5   \\ \hline
\end{tabular}
\end{adjustbox}
\label{tab:exp_rellis}
\end{table*}

\begin{table*}[th!]
\caption{Quantitative comparison of freespace detection on the ORFD dataset}
\centering
\begin{adjustbox}{width=0.95\linewidth}
\begin{tabular}{|c|cc|ccccc|}
\hline
Method  & Supervision     & Modality   & Accuracy (\%) & Precision (\%) & Recall (\%)  & F-score (\%)  & IoU (\%)      \\ \hline \hline
OFF-Net~\cite{min2022orfd} 
        & Supervised      & RGB+SN     & 96.1          & 94.5          & 91.1          & 92.8          & 86.6          \\ \hline
WayFAST~\cite{gasparino2022wayfast}
        & Self-supervised & RGB+D      & 67.1          & 44.6          & \textbf{78.2} & 56.8          & 39.6          \\
Ours    & Self-supervised & RGB+D      & \textbf{85.6} & \textbf{76.8} & 68.7          & \textbf{72.5} & \textbf{56.9} \\ \hline
\end{tabular}
\end{adjustbox}
\label{tab:exp_orfd}
\vspace*{-0.5cm}
\end{table*}

\subsection{Experiments on Path Planning}

\subsubsection{Experimental setting}

We evaluated the traversability estimation methods within the context of path planning on the RELLIS-3D dataset. Each method predicted a local cost map for each frame, and these local maps were accumulated to generate a global cost map. Here, a low cost indicates that the ground can be easily traversed by the robot, whereas a high cost indicates the opposite. Subsequently, paths were generated from the global cost map through the RRT* algorithm~\cite{karaman2011sampling}. The runtime was measured using an NVIDIA GeForce GTX 2080 Ti with an Intel Core i7-9700 K CPU @ 3.60 GHz.

We used a simulator to evaluate the suitability of the predicted global cost map for actual robot navigation. We generated 10 paths for each cost map, and tested whether the robot could actually follow these paths in a simulator. In the simulator, a virtual environment was implemented by converting the point cloud map to a mesh using Poisson surface reconstruction and integrating with the PyBullet physics engine~\cite{coumans2021}.

The following methods were used in the experiment for comparison:
\begin{itemize}
\item WayFAST~\cite{gasparino2022wayfast}: WayFAST uses RGB-D images as input and learns traversability through self-supervision from the driving experience.
\item BEVNet~\cite{shaban2022semantic}: 
BEVNet utilizes a neural network and supervised learning to predict geometric and semantic information from point clouds over time. It classifies terrain into four cost classes to generate traversability maps.
\item GANav~\cite{guan2022ga}: 
GANav is a supervised learning-based method that predicts semantic classes from RGB images and generates traversability maps through semantic class to traversability mapping.
\item Agishev et al.~\cite{agishev2022trajectory}: 
Agishev et al. proposed a learning-based robot-terrain interaction model using RGB images and point clouds. The model leverages the robot’s inclination limits to predict traversability maps.
\end{itemize}

\subsubsection{Experimental results}

The quantitative evaluation results are presented in Tab.~\ref{tab:exp_rellis}. The success rate (SR) represents the ratio of successful driving trials out of 30 trials for each path. Cross-track error (CTE) is the difference between the position of the robot and the desired path while driving along the generated path. Finally, Hausdorff distance (HD)~\cite{taha2015efficient} was used to evaluate the similarity between the GT path and the generated path. For all of these metrics except SR, lower values indicate better performance. In the table, the symbol ‘-’ indicates that a path cannot be generated from the predicted cost map.

WayFAST showed the second-best performance after our method. However, it failed to effectively integrate geometric and visual information during the estimation process, leading to low cost estimates in hazardous areas. This resulted in a failure of the driving trial at Route 1. BEVNet, which relies solely on point clouds for traversability estimation, generated multiple impractical paths that cannot be followed in route 1 and 3. Similarly, GANav, which utilizes only RGB images, classified the puddle as a non-traversable space, through which the robot can easily pass; thus, a traversable space could not be secured and path generation failed in routes 2 and 3. Lastly, Agishev et al. also exhibited low performance, particularly in generating several impractical paths to follow. Our method outperforms the other methods for all of the metrics. This result indicates, in quantitative terms, that our fusion of geometric and visual cues was successful and that our method effectively learned the properties of the robot’s footprint. Furthermore, our method demonstrated the second fastest runtime after WayFAST, demonstrating its feasibility for real-time deployment in actual robots.

The qualitative comparison result is shown in Fig.~\ref{fig:exp_rellis_path}. Our method ensured safety by taking a detour near non-traversable areas (marked in red). In contrast, other methods generated unsafe paths that pass through these areas.

\subsection{Experiments on Freespace Detection}

Here, we evaluated the traversability estimation methods within the context of freespace detection on the ORFD dataset. We classified areas in the predicted cost map as freespace if their cost was less than 0.5. The method used for comparison is WayFAST and OFF-Net~\cite{min2022orfd}. OFF-Net uses RGB and surface normals with human-supervised labels for supervised learning in freespace detection. For quantitative evaluation, we used common metrics in freespace detection~\cite{min2022orfd}, including accuracy, precision, recall, F-score, and intersection of union (IoU). The evaluation results are presented in Tab.~\ref{tab:exp_orfd}.

Our method predicted a high cost for the boundaries of freespace, considering them as dangerous for navigation. This results in narrow freespace predictions, leading to a low recall value. However, avoiding these areas is crucial in real-world driving scenarios.

Overall, the performance of our self-supervised method is comparable to the supervised method, OFF-Net, and superior to another self-supervised method, WayFAST. This result demonstrates the validity of our method in pixel-level evaluation for the freespace detection. 
Additionally, it highlights the strength of our method in off-road environments where generating human-annotated ground truth is challenging.

\section{CONCLUSIONS}

In this study, we addressed the off-road traversability estimation problem by considering surface slope, semantic information, and robot platform as essential factors. We introduced a novel approach that includes the \textit{guide filter network}, integrating surface and semantic information, and the \textit{footprint supervision module}, for self-supervised learning of robot-dependent traversability. Our optimized solution ensures safe robot navigation in diverse terrain and weather conditions, and is compatible with various robot platforms.

\newpage









\bibliographystyle{IEEEtran}
\bibliography{egbib}
\newpage

\end{document}